\documentclass[11pt]{article}
\usepackage{caption}

\usepackage{acl}

\usepackage{times}
\usepackage{latexsym}
\usepackage{multirow}
\usepackage{arydshln}
\usepackage{amsmath}
\usepackage[most]{tcolorbox}
\usepackage{xkeyval}
\usepackage{fvextra}

\usepackage{graphicx}
\usepackage{amsmath}
\usepackage{amssymb}
\usepackage{svg}
\usepackage{fancyvrb}

\usepackage{booktabs}   % \toprule, \midrule, \bottomrule, \cmidrule
\usepackage{graphicx}   % \resizebox (only if the table overflows)

\usepackage[T1]{fontenc}
\usepackage[utf8]{inputenc}

\usepackage{microtype}

\usepackage{inconsolata}

\usepackage{graphicx}

\title{Beyond Verdicts: A Graph-Based Analysis of Human and LLM Reasoning in Scientific Fact-Checking}

\author{Abdul Ghafoor,  Muhammad Arslan Manzoor, Yufang Hou \\
  Interdisciplinary Transformation University Austria (ITU) \\
  \texttt{\{abdul.ghafoor, muhammad.manzoor, yufang.hou\}@it-u.at}
}

\begin{document}
\maketitle
\begin{abstract}
%Misleading scientific claims often cite credible studies while misrepresenting what those studies report. 
%Misleading or false scientific claims on social media sometimes cite legitimate papers while misrepresenting what those papers actually report.
%Existing fact-checking systems can identify whether %an LLM 
%an automatic fact-checking system that also generates explanations
%assigns the \textit{Incorrect} verdict, but they do not show whether the model reaches that verdict through the same reasoning path as human experts, or by a different but still valid reasoning path.
Misinformation that cites legitimate papers can be especially harmful when it distorts what those studies actually report. While existing automatic fact-checking systems based on large language models (LLMs) can assess whether a model assigns an \textit{Incorrect} verdict and can generate explanations for that decision, they typically do not indicate whether the model follows the same reasoning path as human experts or arrives at the verdict through a different but still valid path.
%In this work, we introduce a graph-based framework for comparing human and LLM reasoning paths in scientific fact-checking. 
%Building on a fallacious reasoning framework to explain  real-world misinformation  that misrepresents biomedical
%publications \cite{glockner-etal-2025-grounding},   we represent each explanation as a reasoning graph connecting the \emph{false claim}, \emph{study context}, \emph{study findings}, \emph{fallacy-supporting premises}, and \emph{fallacy labels}. 
In this work, we introduce a graph-based framework (\emph{typed reasoning graph}) for comparing human and LLM reasoning paths in scientific fact-checking. Building on prior work on fallacious reasoning in biomedical misinformation, \textsc{MissciPlus} \cite{glockner-etal-2025-grounding}, we model each explanation as a reasoning graph that links the \emph{false claim} to the relevant \emph{study context}, \emph{study findings}, \emph{fallacy-supporting premises}, and \emph{fallacy labels}. 
This representation enables one-to-one alignment of human and LLM reasoning at the level of fallacy-specific sub-graphs. 
For non-human-aligned LLM paths, we validate grounding in the cited study, relevance to the claim, and sufficiency for the verdict. Using 84 false claims from \textsc{MissciPlus}, we evaluate GPT-5, Claude Opus 4.7, and Qwen3-32B across prompt and evidence settings. Results show distinct performance dimensions: Qwen3-32B has the lowest verdict failure rate, GPT-5 the highest human alignment, and Claude Opus 4.7 weak verdict prediction but often valid reasoning in successful cases.\footnote{Code and data will be released upon acceptance.}

%Using 84 false claims from \textsc{MissciPlus}, we evaluate GPT-5, Claude Opus 4.7, and Qwen3-32B across prompt and evidence settings. 
%Our results show that verdict accuracy and reasoning quality diverge: Qwen3-32B performs best at verdict prediction, GPT-5 aligns most often with human reasoning, and many non-aligned paths remain valid after validation.\footnote{Code and data will be released upon acceptance.}
\end{abstract}

\section{Introduction}

Misinformation is often most persuasive when it cites real scientific studies while misrepresenting what they report \citep{glockner-etal-2024-missci,glockner-etal-2025-grounding}. 
For scientific fact-checking, the main challenge is therefore not only to determine a claim as true or false, but also to assess how the cited evidence is interpreted and how the reasoning from study findings to verdict is constructed.
This distinction is important in the context of fact-checking systems based on large language models (LLMs): 
%a model can produce a correct verdict and a fluent explanation, while its reasoning may not be consistent with human experts or remain grounded in the cited study 
a model may produce a correct verdict and a fluent explanation, yet rely on reasoning that diverges from human experts, is insufficiently grounded in the cited study, or fails to justify the verdict \citep{atanasova-etal-2023-faithfulness,parcalabescu-frank-2024-measuring}.

\begin{figure*}[t]
  \centering
\includegraphics[width=\textwidth]{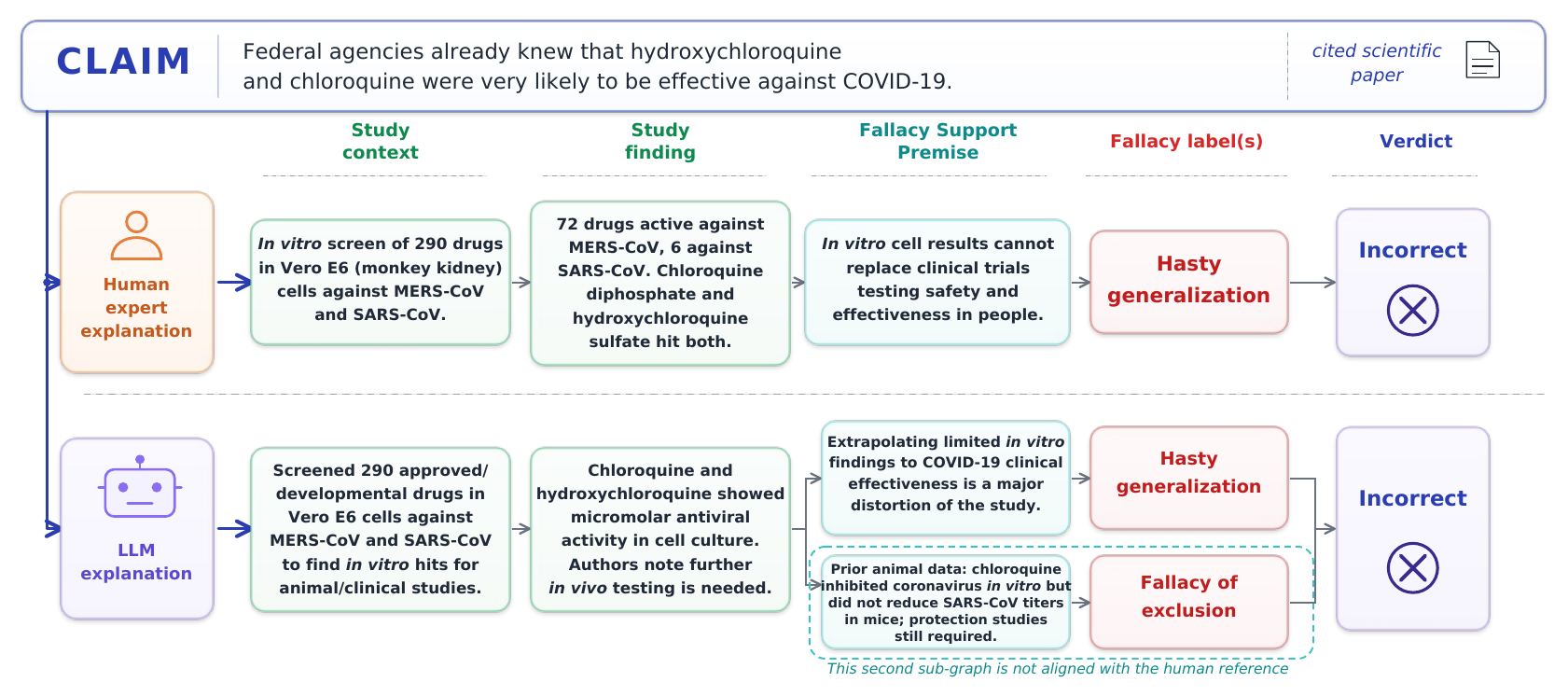}
  \caption{Example reasoning-path comparison for a misleading scientific claim: Both explanations reach the same verdict, \textit{Incorrect}. The LLM aligns with the human reference on the upper reasoning path labeling it as \textit{Hasty Generalization}. In addition, the LLM extracts a second reasoning path labeling it as \textit{Fallacy of Exclusion}}
  \label{fig:intro_diagram}
\end{figure*}

Previous work on scientific and biomedical fact-checking has advanced claim verification, evidence retrieval, rationale selection, and explanation generation \citep{wadden-etal-2020-fact,wadden-etal-2022-scifact,kotonya-toni-2020-explainable-automated,sarrouti-etal-2021-evidence-based,vladika-etal-2024-healthfc}. 
A closely related line of work shows that false scientific claims frequently arise from fallacious reasoning over genuine studies: \textsc{Missci} \citep{glockner-etal-2024-missci} reported such cases as fallacious arguments, and \textsc{MissciPlus} \citep{glockner-etal-2025-grounding} grounds those fallacies in passages from misrepresented studies. 
However, existing evaluations primarily assess verdicts, retrieved evidence, fallacy labels, or free-text explanations, rather than directly comparing whether LLMs and human experts reach the verdict through the same reasoning path. They also leave open a crucial question: when an LLM’s reasoning diverges from expert reasoning, is it still a valid alternative path, or is it ungrounded, irrelevant, or insufficient to justify the verdict?

We frame this gap as a \emph{process-level evaluation} problem: to assess whether an LLM explanation is trustworthy, we must evaluate the inferential structure that connects the cited study to the verdict. 
We define a \emph{typed reasoning graph} as this structure: it links the claim, relevant study context and findings, the premises that expose the misrepresentation, and the fallacy label that characterizes it (Section \ref{sec:methodology}).
%A reasoning path captures this structure by linking the claim, relevant study context and findings, the premises that  expose the misrepresentation,  
%expose a fallacy, 
%and the fallacy label 
%and the fallacy label that characterizes it. 

%Two reasoning paths can yield the same verdict while relying on different evidence, premises, or fallacy types.
%Free-text comparison can obscure these differences.
As illustrated in Figure \ref{fig:intro_diagram}, this structure matters because two explanations can reach the same \textit{Incorrect} verdict %while relying on different evidence, premises, or fallacy types.
while citing different parts of the study, drawing different intermediate premises, or identifying different fallacy types. 
%Figure~\ref{fig:intro_diagram} illustrates one such case, 
%where the human expert and the LLM reach the same \textit{Incorrect} verdict but the LLM extracts an additional fallacy-supporting premise that produces a second, non-aligned fallacy label.
In the example, the human expert and the LLM agree on the final verdict, but the LLM extracts an additional fallacy-supporting premise, leading to a second fallacy label that does not align with the human reference.
Such differences are difficult to capture through verdict accuracy, fallacy labels, or free-text comparison alone.
%We therefore represent each explanation as a \emph{typed reasoning graph}, where nodes encode the functional components of the explanation (i.e., the claim, study context, study findings, fallacy-supporting premises, and fallacy labels) and edges encode the semantic relations among them.
This representation allows us to align human and LLM explanations at the level of \emph{fallacy-specific subgraphs}: fine-grained reasoning units that capture a particular fallacy judgment together with its supporting premises.

%As illustrated in Figure ~\ref{fig:intro_diagram}, we instead represent each explanation as a \emph{typed reasoning graph} and align human and LLM explanations at the level of \emph{fallacy-specific sub-graphs}---a sub-claim unit fine enough to expose where reasoning diverges even when the final verdict agrees.

%On the other side, when an LLM reasoning path diverges from the human reference, divergence is not by itself a failure: an alternative path may still be valid. 
Conversely, divergence from the human reference should not automatically be treated as an error: an LLM may reach the same verdict through a different but still valid reasoning path.
We therefore assess non-aligned reasoning paths by asking whether they are grounded in the cited study, relevant to the claim, and sufficient to justify the verdict. 
This separates principled alternative reasoning from explanations that merely sound plausible, drawing on work in attribution, grounded explanation, and structured rationales \citep{deyoung-etal-2020-eraser,dalvi-etal-2021-explaining,fabbri-etal-2022-qafacteval,min-etal-2023-factscore,li-etal-2024-attributionbench,tang-etal-2024-minicheck}.

%Building on top of our proposed \emph{typed reasoning graph} representation, we %formalize this evaluation through 
%study 
%three research questions:
Building on our \emph{typed reasoning graph} representation, we study three questions that separate verdict correctness from reasoning quality.
(\textbf{RQ1}) Given a misleading scientific claim and its cited study, how reliably do LLMs predict the expected \textit{Incorrect} verdict? (\textbf{RQ2}) When LLMs reach the expected verdict, how often do their reasoning paths align with the human expert path at the level of fallacy-specific sub-graphs? (\textbf{RQ3}) When LLM reasoning is not human-aligned, does it remain grounded in the cited study, relevant to the claim, and sufficient to justify the verdict?

Using 84 false claims and the corresponding scientific papers from \textsc{MissciPlus} \cite{glockner-etal-2025-grounding}, we evaluate GPT-5, Claude Opus 4.7, and Qwen3-32B across prompt and evidence settings. Our results show that verdict accuracy alone substantially overstates reasoning quality (Section \ref{sec:results}). Even under the full-study evidence setting with the detailed prompt, human-aligned reasoning accounts for only a minority of claims: 32.1\% for GPT-5, 8.3\% for Claude Opus 4.7, and 15.5\% for Qwen3-32B. However, many non-aligned reasoning paths are still valid after validation, indicating that LLMs often reach correct verdicts through alternative reasoning paths rather than by reproducing expert rationales. This pattern is most pronounced for Qwen3-32B, which has the lowest verdict failure rate but limited human alignment, and for Claude Opus 4.7, whose high verdict failure rate contrasts with a substantial share of valid non-aligned reasoning among the remaining cases. 

These findings suggest that LLM-based systems may support real-world scientific fact-checking by generating structured reasoning paths for inspection. Experts can use these paths as candidate explanations to verify or revise, while the 
% general 
public can use them to see how a verdict is justified and where a claim misrepresents cited evidence. However, such systems should remain explanation aids rather than standalone arbiters, since assessing grounding and sufficiency often requires domain expertise.

To summarize, our main contributions are:
(1) we propose a graph-based framework for comparing human and LLM reasoning paths in scientific fact-checking that decomposes each explanation into \emph{fallacy-specific sub-graphs} and aligns them one-to-one, moving beyond final-verdict evaluation; 
(2) we introduce a validation for non-human-aligned sub-graphs that define valid alternative reasoning paths from paths that are not grounded in the cited study or inadequate to justify the verdict; (3) we apply this method to GPT-5, Claude Opus 4.7, and Qwen3-32B on 84 false claims from \textsc{MissciPlus}, and show that verdict accuracy and reasoning-path quality dissociate: the model with the highest \textit{Incorrect} recall is not the model with the highest human-aligned reasoning rate.

\begin{figure*}[t]
    \centering
    \includegraphics[width=\textwidth]{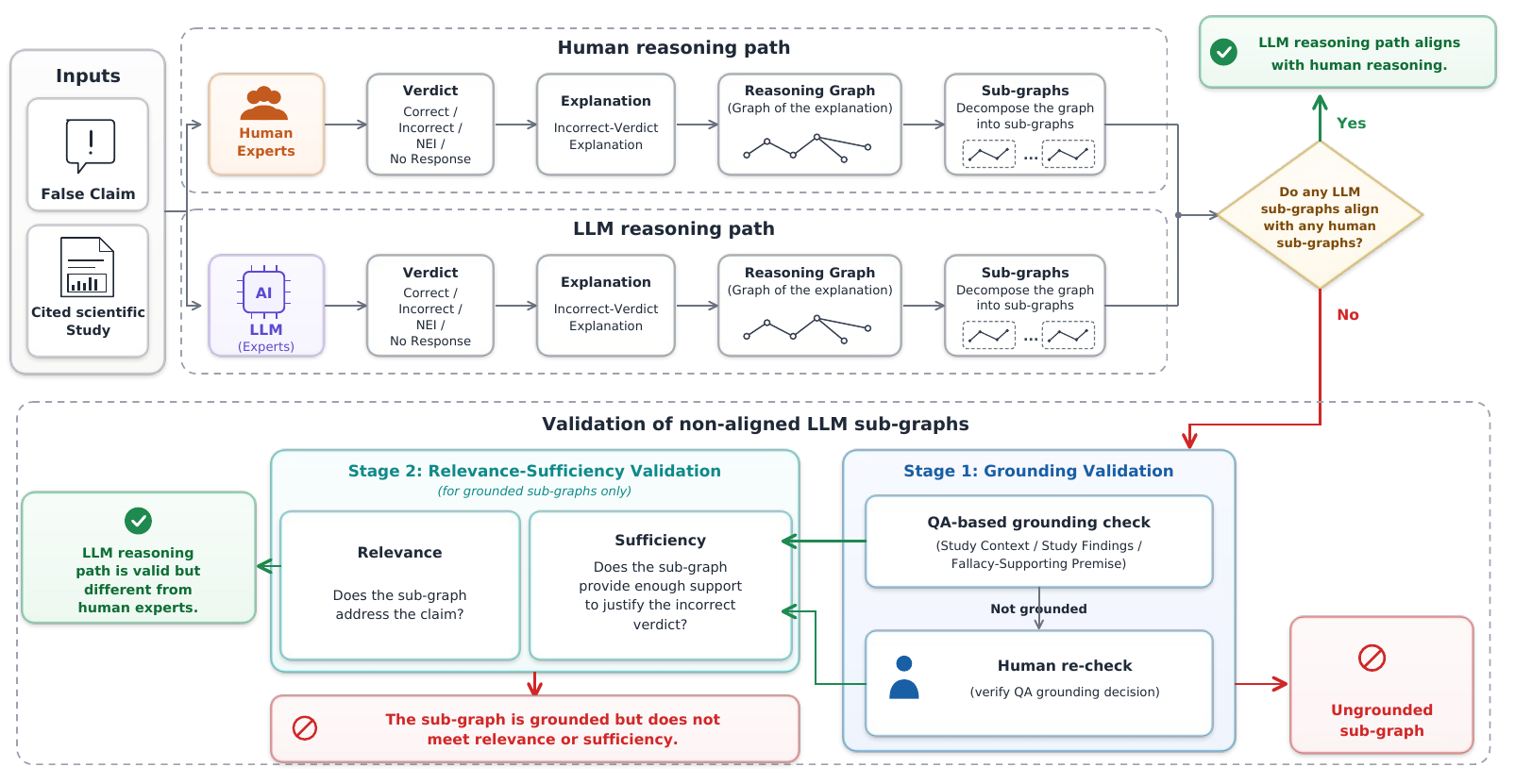}
    \caption{Overview of our graph-based framework for comparing human and LLM reasoning in scientific fact-checking. For \textit{Incorrect}-verdict cases, human and LLM explanations are converted into reasoning graphs and compared at the level of fallacy-specific sub-graphs. Non-aligned LLM reasoning paths are further evaluated for study grounding, relevance to the claim, and sufficiency for the verdict.}
    \label{fig:methodology}
\end{figure*}

\section{Problem Definition and Method}
\label{sec:methodology}
We study whether LLMs follow the same evidence-based reasoning paths as human expert fact-checkers when evaluating scientific claims that misrepresent cited studies. We approach Human-LLM reasoning comparison as a graph alignment and validation problem.  Given an inaccurate scientific claim and its cited study, we generate LLM verdicts and explanations, convert \textit{Incorrect}-verdict explanations into reasoning graphs, and compare them with human expert reasoning graphs.

As shown in Figure \ref {fig:methodology}, proposed framework proceeds in four stages. 
%First, we compute verdict distributions across repeated LLM seeds. 
First, we characterize model verdict behavior by aggregating outputs across repeated runs. 
Second, for outputs that assign the expected \textit{Incorrect} verdict, we represent both human and LLM explanations using a shared reasoning-graph schema (Section \ref{subsec:reasoning-graph-construction}). 
Third, we compare both reasoning using fallacy-specific sub-graphs (Section~\ref{subsec:subgraph-alignment}). Fourth, non-aligned LLM sub-graphs are further assessed for study grounding, relevance, and sufficiency (Section \ref{subsec:validation}).

\subsection{Task Formulation}
\label{subsec:task-formulation}

%Each instance consists of an inaccurate scientific claim $c$ and a cited primary study $s$. 
%The claim misrepresents the study by overstating, generalizing, or incorrectly interpreting the evidence reported in $s$. 
%Given $c$ and evidence $e$ from the cited study, an LLM produces a verdict $v$ and a natural-language explanation $x$.

Each instance consists of a misleading scientific claim $c$ and a cited primary study $s$. The claim misrepresents the study by overstating, generalizing, or incorrectly interpreting the evidence reported in $s$. Given $c$ and evidence $e$ from the cited study, we define the human reference as $(v, x)$, where $v$ is the expert verdict and $x$ is the expert explanation; an LLM produces a corresponding output $(v', x')$, where $v'$ is the model verdict and $x'$ is its natural-language explanation.

The model is prompted to output a verdict in \{\textit{Correct}, \textit{Incorrect}, \textit{Not Enough Information}\}; outputs that are missing, malformed, or refused are labeled \textit{No Response} for analysis. The analysis-time verdict space is therefore \{\textit{Correct}, \textit{Incorrect}, \textit{Not Enough Information}, \textit{No Response}\}.
Since all claims in our dataset are false or misleading, the expected verdict is \textit{Incorrect}. 

We treat verdict prediction, reasoning-path alignment, and validation of non-aligned reasoning as distinct evaluation targets. Verdict prediction asks whether the model reaches the expected \textit{Incorrect} verdict. Reasoning-path alignment asks, among expected-verdict outputs, whether the LLM follows the same fallacy-specific reasoning path as the human expert. Validation of non-aligned reasoning asks whether divergent LLM paths are nevertheless grounded in the cited study $s$, relevant to the claim $c$, and sufficient to justify the verdict.

\subsection{Reasoning Graph Construction}
\label{subsec:reasoning-graph-construction}

Building on prior work on fallacious reasoning in biomedical misinformation \cite{glockner-etal-2025-grounding}, we represent both human and LLM explanations ($x$ and $x'$) using a shared reasoning-graph schema. 
The graph makes explicit the reasoning chain by which the cited study is used to identify the mismatch between the claim and the evidence, and how this mismatch supports a fallacy assignment. Each graph follows the structure:

\textbf{Study Context: } captures information needed to interpret the cited study, such as the study design, population, scope, setting, methodology, or limitations. 
\textbf{Study Findings: } capture the relevant reported outcomes, conclusions, effects, or empirical observations. 
\textbf{Fallacy-Supporting Premises: } are reasoning statements that justify the assignment of a fallacy label by explaining the mismatch between the claim and the cited evidence. 
\textbf{Fallacies: } are the reasoning-error labels assigned to the claim.

Formally, each reasoning graph is a directed graph $G=(V,E)$ whose nodes $V$ are typed by the five classes above (\textit{Claim}, \textit{Study Context}, \textit{Study Findings}, \textit{Fallacy-Supporting Premises}, \textit{Fallacies}) and whose edges $E$ follow the chain order shown.

We use GPT-5 as the graph constructor. Given a free-text explanation $x$ or $x'$, GPT-5 is prompted to extract the \textit{Study Context}, \textit{Study Findings}, and \textit{Fallacy-Supporting Premises} fields with spans \emph{copied verbatim} from the explanation (paraphrasing, summarisation, and rewording are explicitly disallowed) and to assign each premise a \textit{Fallacy} label drawn from the nine-class taxonomy defined in \textsc{MISSCI} \cite{glockner-etal-2024-missci}. 
%The output is returned as JSON conforming to a fixed schema; 
The full prompt is provided in Appendix Figure~\ref{fig:graph-extraction-prompt}. To calibrate our LLM-based graph constructor, we conducted human annotations of the constructed reasoning graphs. Appendix Figure~\ref{fig:human_vs_gpt5_alignment} provides details. The graph constructor achieved strong alignment with human annotators, indicating that it reliably captures the reasoning structure expressed in LLM-generated or human-written explanations.
% For the GPT-5-generated explanations, we also conducted human annotation to convert the explanation text into reasoning graphs. Appendix Figure~\ref{fig:human_vs_gpt5_alignment} presents the alignment between the LLM and human annotator in constructing these reasoning graphs.

\subsubsection{Fallacy-Specific Sub-graphs:}
A single explanation may contain multiple reasoning errors. 
We therefore decompose each reasoning graph into fallacy-specific sub-graphs and use them as the basic unit of alignment and validation. 
Each sub-graph is centered on one fallacy type and contains the study context, study findings, and fallacy support premise. 

% used to justify the assignment of that fallacy label.

\subsection{Human and LLM Sub-graph Alignment}
\label{subsec:subgraph-alignment}
For each claim, we compare every LLM fallacy-specific subgraph against human reference. An LLM subgraph is marked as human-aligned if it matches at least one human subgraph. A match requires two conditions: the two subgraphs must assign the same fallacy type, and their text-based components must be semantically aligned.

We assess semantic alignment with three LLM judges: GPT-5, Qwen3-32B, and Claude Opus 4.7. Each judge assigns a semantic textual similarity (STS) score from 0 to 5 to each component pair, following the rubric in Appendix Figure~\ref{fig:sts-prompt}. A score of 5 indicates complete semantic equivalence, whereas a score of 0 indicates unrelated or contradictory meanings. We aggregate the three judge scores by majority vote; if no two judges assign the same score, we use the median.

A component pair is considered matched if its final STS score is at least 3. This threshold corresponds to cases where the two components express the same semantic facet, although some details may differ or be omitted. 

% An LLM subgraph is therefore considered human-aligned only if it has the same fallacy type as a human subgraph and all three text-based components meet the STS threshold.

%We compare each LLM sub-graph with the human sub-graphs for the same claim. 
%An LLM sub-graph is marked as human-aligned if it matches at least one human sub-graph. 
%A match requires two conditions.

%First, the LLM and human sub-graphs must assign the same fallacy type. 
%Second, their text-based components must be semantically aligned. 

%To assess semantic alignment, three LLM judges---GPT-5, Qwen3-32B, and Claude Opus 4.7---assign a semantic textual similarity (STS) score to each component pair using the 0--5 STS rubric described in Appendix Figure~\ref{fig:sts-prompt}.
%A score of 5 indicates complete semantic equivalence, while a score of 0 indicates different meanings. 
%We aggregate the three judge scores using majority voting. 
%If at least two judges assign the same score, that score is used as the final STS score. 
%If all three judges assign different scores, we use the median score.

%A component pair is considered matched if its final STS score is at least 3. 
%We set the matching threshold to $\mathrm{STS}\geq3$, corresponding to cases where two components describe the same semantic facet, even if some details differ or are omitted. 
%Thus, an LLM sub-graph is considered human-aligned only when the fallacy type matches and all three text-based components satisfy the STS threshold.

\subsection{Validation of Non-Aligned Sub-graphs}
\label{subsec:validation}

Non-alignment with the human reference does not necessarily imply invalid reasoning. 
An LLM may identify a reasoning path that is absent from the human explanation but still grounded in the cited study and sufficient to support the \textit{Incorrect} verdict. 
We therefore validate non-aligned LLM sub-graphs in two stages: grounding and relevance--sufficiency validation.

% \subsubsection{Grounding Validation}
% \label{subsubsec:grounding-validation}
\paragraph{Grounding Validation.}
Grounding validation checks whether each component of a non-aligned sub-graph is supported by the cited study.
We evaluate grounding for \textit{Study Context}, \textit{Study Findings}, and \textit{Fallacy-Supporting Premise}.
For each component, we use GPT-5 to generate five questions capturing its key information and answers each question using two sources: the component itself and the full cited study, refer to Appendix Figure~\ref{fig:prompt-qa-question-gen} for full prompt.
One author of this paper manually compares the component-based and study-based answers.
If any answer pair does not match, the component is initially marked as not grounded.

% To reduce false rejections caused by QA or annotation errors, all initially not-grounded components undergo a second human re-check by the same author.
% If the re-check confirms study support, the component is treated as grounded; otherwise, it remains not grounded.
% A sub-graph is classified as \textit{not grounded} if at least one of its components remains unsupported after re-check.

% \subsubsection{Relevancy and Sufficiency Validation}
% \label{subsubsec:relevance-sufficiency-validation}
\paragraph{Relevancy and Sufficiency Validation.}
Grounded non-aligned sub-graphs are further evaluated for relevance to the claim and sufficiency for the \textit{Incorrect} verdict using the prompt in Appendix Figure~\ref{fig:relevance-sufficiency-prompt}. 
Three LLM judges GPT-5, Claude Opus 4.7, and Qwen3-32B score each sub-graph on both dimensions using a three-point scale: 2 = clearly satisfied, 1 = partially satisfied, and 0 = not satisfied. 
Scores are aggregated using the same majority-vote and median procedure described in Section~\ref{subsec:subgraph-alignment}. Finally, we combine the final relevance and sufficiency scores into a single label. 
A sub-graph receives a combined score of 2 only if both receive 2.

% \subsection{Outcome Categories}
% \label{subsec:outcome-categories}

% After fallacy-specific subgraph alignment  (Section \ref{subsec:subgraph-alignment}) and validation of non-aligned reasoning (Section \ref{subsec:validation}), each LLM sub-graph is assigned to one of four outcome categories. \textbf{(1) Human-aligned reasoning: }
% The LLM sub-graph matches at least one human sub-graph with the same fallacy type and semantically aligned reasoning components.
%  \textbf{(2) Valid but different reasoning path: }
% The LLM sub-graph does not align with the human graph but is grounded, relevant, and sufficient to justify the \textit{Incorrect} verdict. \textbf{(3) Not grounded reasoning: }
% The LLM sub-graph contains at least one component that remains unsupported by the cited evidence after human re-check. \textbf{(4) Grounded but inadequate reasoning: }
% The LLM sub-graph is grounded in the cited evidence but fails the relevance and sufficiency validation.

\section{Experimental Setup}
\label{sec:experimental-setting}

\subsection{Data and Evidence Settings}
\label{subsec:data-evidence}

We use the \textsc{MissciPlus} \citep{glockner-etal-2025-grounding} test set, which contains 84 false scientific claims paired with cited primary studies and claim-relevant study passages. 
Each claim is associated with a HealthFeedback\footnote{\url{https://science.feedback.org/}} expert review explaining why the cited study does not support the claim.

One author of this paper manually extracts, for each of the 84 instances, the portion of the HealthFeedback review that explains the mismatch between the claim and the cited study.
The extracted text serves as the human expert reasoning reference for the \textit{Incorrect} verdict. We treat HealthFeedback reviewers as the underlying experts and the extracted spans as a faithful selection of their reasoning; we do not paraphrase, summarise, or otherwise rewrite the review text.

We consider two evidence settings. 
In the selected-passage setting, the model receives the claim-relevant passages provided in \textsc{MissciPlus} \citep{glockner-etal-2025-grounding} as the evidence. 
In the full-study setting, the model receives the full text of the cited primary study as the evidence. 

\subsection{LLM Verdict and Explanation Generation}
\label{subsec:llm-generation}

For each claim -- evidence pair, we prompt an testing LLM to assign a verdict and generate an explanation.\footnote{The model outputs a verdict in \{\textit{Correct}, \textit{Incorrect}, \textit{Not Enough Information}\} and a free-text explanation. 
Outputs without a valid verdict or usable response are labeled \textit{No Response} for analysis. }  
Each experimental configuration is defined by the model, prompt template, and evidence setting. 
We use two prompt templates: a concise prompt adapted from \citet{glockner-etal-2025-grounding} and a detailed fact-checking prompt designed for this study.
The detailed prompt is provided in Appendix Figure~\ref{fig:factcheck-prompt}.
For each configuration, we run the same input three times to account for output variability. 

For claim $i$ and verdict label $\ell$, we compute the per-claim verdict proportion as:
\begin{equation}
p_{i,\ell} = \frac{n_{i,\ell}}{3},
\end{equation}

\noindent
where $n_{i,\ell}$ is the number of runs, out of three, in which the model produces label $\ell$ for claim $i$. 
We then average these per-claim proportions across all $N$ claims:
\begin{equation}
\bar{p}_{\ell} = 100 \cdot \frac{1}{N} \sum_{i=1}^{N} p_{i,\ell}.
\end{equation}

\noindent
The resulting value $\bar{p}_{\ell}$ represents the overall percentage assigned to verdict label $\ell$ under a given experimental configuration. 
We also report verdict consistency, defined as the proportion of claims for which all three runs produce the same verdict.

%Verdict distributions are computed across the three runs for each model -- prompt (2 variations) -- evidence configuration (2 variations).
Verdict distributions are computed over three runs for each experimental configuration, defined by model, one of two prompt templates, and one of two evidence settings.
%following Section~\ref{subsec:llm-generation}. 
%The same three LLMs are used as independent judges for relevance--sufficiency validation, with aggregation performed as described in Section~\ref{subsubsec:relevance-sufficiency-validation}.
Since all claims are false or misleading, the expected verdict for each instance is \textit{Incorrect}. The reported \textit{Incorrect} rate therefore corresponds to recall on the expected label rather than overall accuracy; the \textit{Correct} column represents Type-I errors (false-positive support of misleading claims), and the \textit{NEI} column represents abstention. We discuss this asymmetry and its implications in Section~\ref{subsec:discussion}.

% Each configuration is run three times to account for output variability. 
% Temperature is set to 0.1 for all models.

\subsection{Reasoning Graph Construction Settings}
For reasoning-graph analysis, we focus on the full-study evidence with the detailed prompt. This setting gives the model the strongest opportunity to ground its explanation in the complete cited study while following explicit fact-checking instructions. Because each claim-model pair is run three times, we select one valid explanation that assigns the expected \textit{Incorrect} verdict for graph construction. We then convert the selected explanations into reasoning graphs and evaluate them using the alignment and validation procedures described in Sections~\ref{subsec:reasoning-graph-construction}

% For reasoning-graph construction, we use one valid \textit{Incorrect}-verdict explanation per claim--model pair from the full-study/detailed-prompt configuration.

% Reasoning graphs are constructed from \textit{Incorrect}-verdict explanations and evaluated using the alignment and validation procedures described in Sections~\ref{subsec:reasoning-graph-construction}--\ref{subsec:outcome-categories}. 

% For the reasoning-graph analysis, we use the full-study evidence setting with the detailed prompt. 
% This setting gives the model the strongest opportunity to ground its explanation in the complete cited study while following explicit fact-checking instructions.

%\subsection{Dataset}

%We evaluate on the 84 MISSCIPlus test instances described in Section~\ref{subsec:data-evidence}.

\subsection{Testing Models}

We evaluate three LLMs as fact-checkers: GPT-5 \citep{DBLP:journals/corr/abs-2601-03267}, Claude Opus 4.7\citep{anthropic2026claudeopus47}, and Qwen3-32B \citep{DBLP:journals/corr/abs-2505-09388}. 
GPT-5 is accessed through the Academic-AI gateway, Claude Opus 4.7 through the Anthropic API, and Qwen3-32B is served locally using vLLM. 
Within each experimental configuration, all models receive the same claim, evidence, and prompt template.
Temperature is set to 0.1 for all models.

\section{Results and Discussion}
\label{sec:results}

We evaluate models along two complementary axes: verdict prediction and reasoning-path quality. 
The first measures whether models identify false claims as \textit{Incorrect}; the second examines whether their explanations align with human expert reasoning or, when non-aligned, still form grounded and sufficient alternative reasoning paths.

\subsection{Verdict Prediction}
\label{subsec:verdict-results}

Table~\ref{tab:combined_results} reports verdict distributions across models, evidence settings, and prompt templates.
Since all instances in the dataset are false claims, higher \textit{Incorrect} rates correspond to higher recall of the expected label.
Overall, the detailed prompt improves \textit{Incorrect} prediction for GPT-5 and Qwen3-32B. 
For GPT-5, the \textit{Incorrect} rate increases from 32.54\% to 74.60\% in the full-study setting and from 42.46\% to 69.05\% in the selected-passage setting. 
Qwen3-32B achieves the highest overall \textit{Incorrect} rate, reaching 76.98\% with selected passages and the detailed prompt, and 73.41\% with full-study evidence and the detailed prompt. 
These results suggest that explicit fact-checking instructions help models connect the claim to the cited evidence and produce the expected verdict.

The effect of evidence length is model-dependent. 
GPT-5 performs best with full-study evidence under the detailed prompt, whereas Qwen3-32B performs best with selected passages. 
This indicates that more evidence is not uniformly beneficial: selected passages may reduce distracting context for some models, while full studies may help others access information needed for a more complete interpretation. 
Surprisingly, Claude Opus 4.7 shows lower \textit{Incorrect} rates overall, with its best result reaching 44.58\% in the full-study/detailed-prompt setting. 
However, Claude is highly stable across repeated runs, reaching 100\% consistency in both detailed-prompt settings. 
This contrast shows that %consistency should not be interpreted as correctness; 
a model can be stable while still assigning many false claims to the wrong verdict label.

\begin{table*}[t]
\centering
\small
\setlength{\tabcolsep}{4pt}
\begin{tabular}{ll l cccc c}
\hline
\textbf{Model} & \textbf{Evidence} & \textbf{Prompt} & \textbf{Correct} & \textbf{Incorrect} & \textbf{NEI} & \textbf{No Response} & \textbf{Consistency} \\
\hline

\multirow{5}{*}{\textbf{Qwen3-32B}} 
& Internal Knowledge (IK)
& IK\_Prompt & 4.76 & \textbf{82.14} & 8.33 & 4.76 & 100.00 \\
& \multirow{2}{*}{Full Study} 
& Short Prompt    & 30.56 & 39.68 & 29.76 & 0.00 & 80.95 \\
& 
& Detailed Prompt & 23.81 & 73.41 & 2.78  & 0.00 & 85.71 \\
& \multirow{2}{*}{Selected Passages} 
& Short Prompt    & 27.78 & 31.75 & 40.48 & 0.00 & 79.76 \\
& 
& Detailed Prompt & 19.84 & 76.98 & 3.17  & 0.00 & 86.90 \\
\hline

\multirow{5}{*}{\textbf{GPT-5}} 
& Internal Knowledge (IK)
& IK\_Prompt & 7.54 & \underline{\textbf{84.92}} & 7.54 & 0.00 & 78.57 \\
& \multirow{2}{*}{Full Study} 
& Short Prompt    & 20.63 & 32.54 & 46.83 & 0.00 & 71.43 \\
& 
& Detailed Prompt & 15.87 & 74.60 & 9.52  & 0.00 & 72.62 \\
& \multirow{2}{*}{Selected Passages} 
& Short Prompt    & 18.65 & 42.46 & 38.10 & 0.79 & 77.38 \\
& 
& Detailed Prompt & 19.05 & 69.05 & 11.90 & 0.00 & 95.24 \\
\hline

\multirow{5}{*}{\textbf{Claude Opus 4.7}} 
& Internal Knowledge (IK)
& IK\_Prompt & 12.70 & \textbf{76.59} & 9.52 & 1.19 & 85.71 \\
& \multirow{2}{*}{Full Study} 
& Short Prompt    & 32.14 & 19.84 & 45.63 & 2.38 & 87.80 \\
& 
& Detailed Prompt & 38.55 & 44.58 & 15.66 & 1.20 & 100.00 \\
& \multirow{2}{*}{Selected Passages} 
& Short Prompt    & 48.81 & 34.52 & 16.67 & 0.00 & 95.24 \\
& 
& Detailed Prompt & 26.59 & 48.02 & 21.83 & 3.57 & 97.62 \\
\hline

\multirow{3}{*}{\textbf{Claude Sonnet 4.6}} 
& Internal Knowledge (IK)
& IK\_Prompt & 8.73 & \textbf{78.97} & 11.11 & 1.19 & 88.10 \\
& Full Study
& Detailed Prompt & 15.48 & 76.19 & 7.14 & 1.19 & -- \\
& Selected Passages
& Detailed Prompt & 18.25 & 66.67 & 12.70 & 2.38 & 94.05 \\
\hline

\end{tabular}
\caption{LLM-based fact-checking results across prompting strategies and evidence settings.
Values report average verdict distribution (\%) over three runs and model consistency (agreement across runs), except Claude Sonnet 4.6 full-study results, which are from seed 1 only. The internal-knowledge setting uses no external evidence.}
\label{tab:combined_results}
\end{table*}

\begin{figure}[t]
  \centering
  \includegraphics[width=\columnwidth]{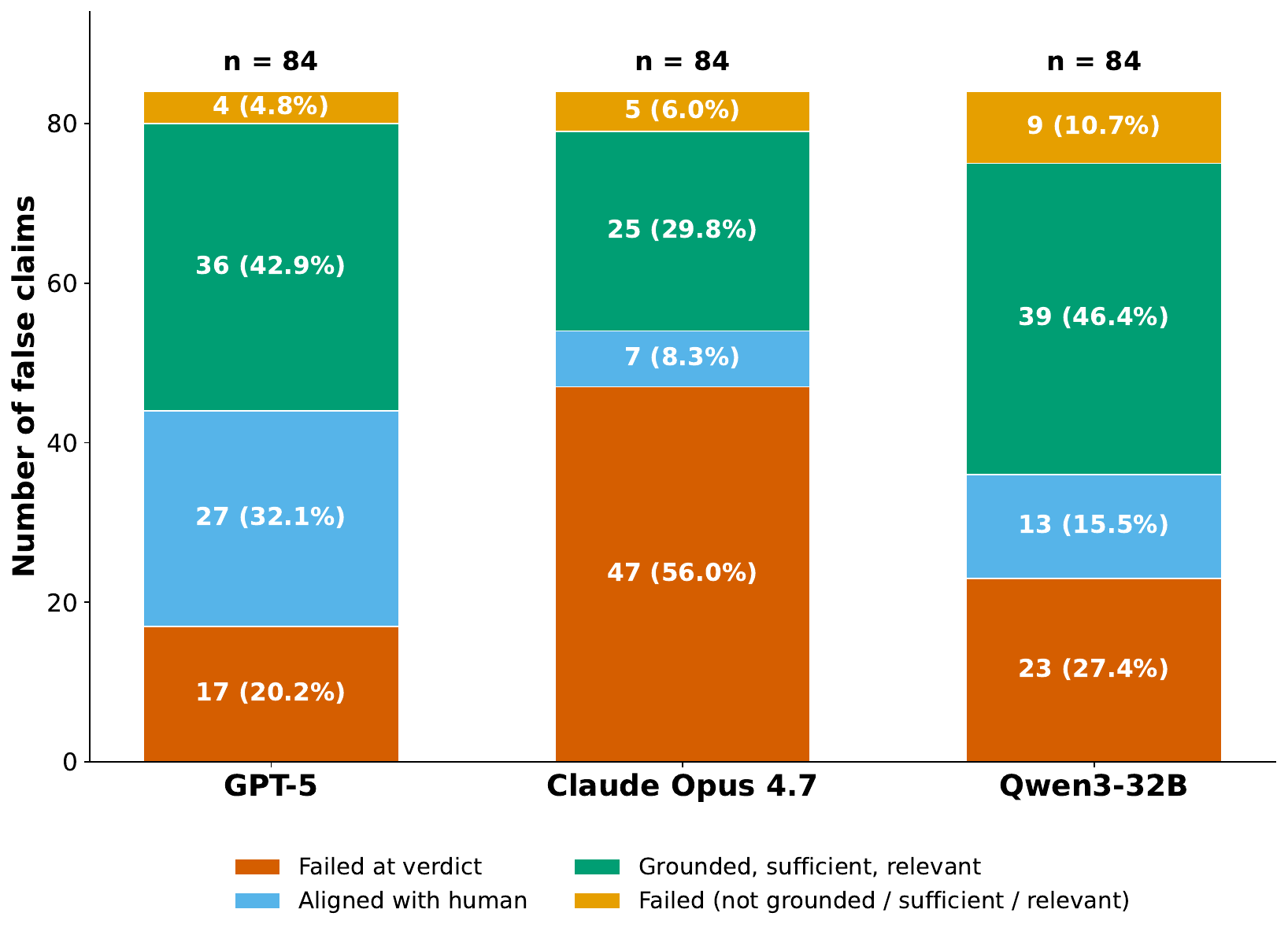}
  \caption{Claim-level outcome breakdown under the full-study evidence setting with the detailed prompt ($n=84$ per model).}
  \label{fig:claim_level_dis}
\end{figure}

\subsection{Reasoning-Path Alignment and Validation}
\label{subsec:reasoning-results}

Table~\ref{tab:alignment_validation} reports the alignment and validation results for LLM reasoning sub-graphs. 
A sub-graph is accepted if it either aligns with a human sub-graph or, when non-aligned, passes grounding and receives the maximum relevance and sufficiency score. 
This evaluation tests whether models reach the expected verdict through expert-like or otherwise valid reasoning paths \citep{deyoung-etal-2020-eraser}. To reduce false rejections caused by QA or annotation errors, all initially not-grounded components undergo a second human re-check by the same author.
If the re-check confirms study support, the component is treated as grounded; otherwise, it remains not grounded. A sub-graph is classified as \textit{not grounded} if at least one of its components remains unsupported after re-check.

GPT-5 produces the strongest reasoning results: 19.0\% of its 158 sub-graphs are human-aligned, and 81.6\% are accepted overall after validation.
This indicates that GPT-5 often reasons differently from the human expert, but many of its alternative paths are grounded, relevant, and sufficient for the \textit{Incorrect} verdict. The hydroxychloroquine case in Figure~\ref{fig:intro_diagram} illustrates this pattern: the human reference assigns only \textit{Hasty generalization}, while the LLM additionally introduces \textit{Fallacy of exclusion} via an extra fallacy-supporting premise. As confirmed by the validation pipeline (Section~\ref{subsec:validation}), this additional subgraph is grounded, relevant, and sufficient to be accepted as a valid alternative reasoning path.
%and the validation pipeline (Section~\ref{subsec:validation}) decides whether that additional sub-graph is grounded, relevant, and sufficient to be accepted as a valid alternative path.
Qwen3-32B shows a different pattern. 
Although it achieves the strongest verdict-level performance in Table~\ref{tab:combined_results}, only 10.2\% of its sub-graphs are human-aligned, with 80.3\% accepted overall. 
Thus, high verdict accuracy does not necessarily imply strong human reasoning alignment.
Claude Opus 4.7 obtains the highest accepted proportion, with 86.9\% of its evaluated sub-graphs accepted, despite its lower \textit{Incorrect} verdict rate. 
This suggests that Claude less often reaches the expected verdict, but when it does, its reasoning is frequently grounded and adequate. Figure~\ref{fig:claim_level_dis} further shows that models differ not only in verdict failures, but also in whether their \textit{Incorrect}-verdict explanations are human-aligned, valid but different, or rejected after validation. 
Overall, these results show that verdict prediction and reasoning quality capture distinct aspects of model behavior.

\begin{table*}[t]
\centering
\scriptsize
\setlength{\tabcolsep}{3pt}
\resizebox{\textwidth}{!}{%
\begin{tabular}{lccccccccccccccc}
\toprule
\textbf{LLM (\textit{n})}
& \textbf{Human-}
& \multicolumn{12}{c}{\textbf{Not Human-Aligned}}
& \multicolumn{2}{c}{\textbf{Final}} \\
& \textbf{Aligned}
& \textbf{Grounded (QA)}
& \textbf{Grounded (final)}
& \textbf{Ungrounded}
& \multicolumn{3}{c}{\textbf{Relevance}}
& \multicolumn{3}{c}{\textbf{Sufficiency}}
& \multicolumn{3}{c}{\textbf{Rel.+Suff.}}
& \textbf{Accepted}
& \textbf{Not Accepted} \\
\cmidrule(lr){2-2}
\cmidrule(lr){3-14}
\cmidrule(lr){15-16}
&
&
&
&
& \textbf{2} & \textbf{1} & \textbf{0}
& \textbf{2} & \textbf{1} & \textbf{0}
& \textbf{2} & \textbf{1} & \textbf{0}
& & \\
\midrule
GPT-5 (158)
& 19.0 & 62.7 & 74.7 & 6.3
& 74.1 & 0.6 & 0.0
& 63.3 & 11.4 & 0.0
& 62.7 & 12.0 & 0.0
& 81.7 & 18.3 \\

Qwen3-32B (127)
& 10.2 & 49.6 & 81.1 & 8.7
& 79.5 & 0.8 & 0.8
& 70.1 & 10.2 & 0.8
& 70.1 & 10.2 & 0.8
& 80.3 & 19.7 \\

Claude Opus 4.7 (84)
& 11.9 & 69.0 & 82.1 & 6.0
& 81.0 & 1.2 & 0.0
& 75.0 & 6.0 & 1.2
& 75.0 & 6.0 & 1.2
& 86.9 & 13.1 \\
\bottomrule
\end{tabular}%
}

\caption{Alignment and validation of LLM reasoning sub-graphs, reported as \% of each model's total sub-graph set. \emph{Human-Aligned} + \emph{Grounded (final)} + \emph{Ungrounded} sum to 100\% per row; \emph{Accepted} + \emph{Not Accepted} also sum to 100\%. \emph{Grounded (QA)} is the subset of \emph{Grounded (final)} admitted by the automatic QA layer alone; QA over-flagged premises as not-grounded due to overly generic questions and granularity mismatches with the document ; human re-verification corrected these cases, raising the \emph{Grounded (final) number}. Relevance, Sufficiency, and Rel.+Suff.\ columns report the three-LLM-judge consensus score distribution (2/1/0) over the \emph{Grounded (final)} pool. A sub-graph is \emph{Accepted} if it is human-aligned, or if its Rel.+Suff.\ score is 2; otherwise \emph{Not Accepted}. A high accepted rate alone does not imply strong overall performance, since Claude performs poorly at verdict prediction.}
\label{tab:alignment_validation}
\end{table*}

\begin{figure}[t]
  \centering
  \includegraphics[width=\columnwidth]{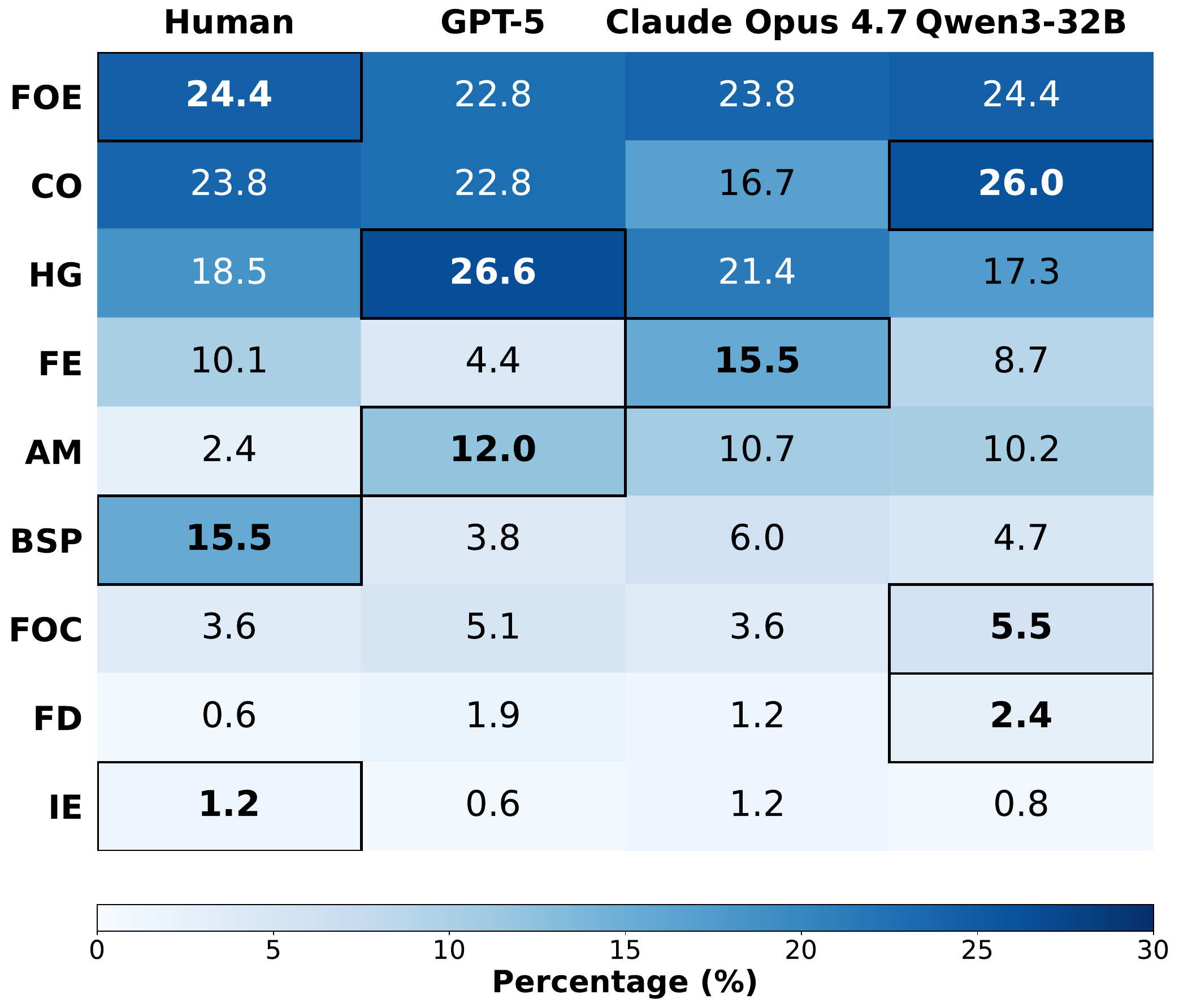}
  \caption{Fallacy-label distributions across human and model reasoning graphs. Black outlines mark the highest percentage within each fallacy type. The most frequent labels are \textit{Fallacy of Exclusion (FOE)}, \textit{Causal Oversimplification (CO)}, and \textit{Hasty Generalization (HG)}. Fallacy abbreviations are listed in Appendix Table~\ref{tab:fallacy-abbrev}.}
  \label{fig:qa_fallacy_dist}
\end{figure}

\begin{table}[t]
\centering
\small
\begin{tabular}{lc}
\toprule
Model pair & Cohen's $\kappa$ \\
\midrule
GPT-5 $\leftrightarrow$ Qwen3-32B       & \textbf{0.511} \\
GPT-5 $\leftrightarrow$ Claude Opus 4.7 & 0.466 \\
Claude Opus 4.7 $\leftrightarrow$ Qwen3-32B & 0.428 \\
\bottomrule
\end{tabular}
\caption{Cross-model agreement in fallacy assignment. Cohen’s $\kappa$ measures whether two models assign the same fallacy label to the same claim, showing moderate agreement across model pairs. }
\label{tab:cross-model-agreement}
\end{table}

Table~\ref{tab:cross-model-agreement} reports moderate cross-model agreement on fallacy choice. 
Agreement is highest between GPT-5 and Qwen3-32B ($\kappa=0.511$), followed by GPT-5 and Claude Opus 4.7 ($\kappa=0.466$), and Claude Opus 4.7 and Qwen3-32B ($\kappa=0.428$). 
This moderate agreement suggests that fallacy assignment remains a challenging and model-sensitive aspect of scientific fact-checking. 
It also supports the need to evaluate reasoning components rather than relying only on final verdicts. 

\subsection{Contamination check}
\label{subsec:contamination-check}
Table~\ref{tab:latest-claims-contamination} reports a small contamination check on 10 recent false claims from 2025--2026 that are not part of the main dataset. 
All three models still identify most of these claims as \textit{Incorrect}, but their human-alignment rates differ substantially: GPT-5 reaches 64.7\% human alignment, compared with 27.8\% for Claude Opus 4.7 and 11.8\% for Qwen3-32B. 
This limited check suggests that the observed reasoning behavior is not confined to the original benchmark instances, and that verdict-level success can again diverge from reasoning alignment.
%Although this check is small, it suggests that the observed reasoning behavior is not limited to the original benchmark instances and that verdict-level success can again diverge from reasoning alignment.

\subsection{Discussion}
\label{subsec:discussion}

Our results clarify the relationship between verdict correctness and reasoning quality in scientific fact-checking. 
First, at the level of verdict prediction (\emph{RQ1}), Table~\ref{tab:combined_results} shows that LLMs often identify misleading scientific claims as \textit{Incorrect}, but their performance depends on the model, prompting strategy, and evidence setting. 
Detailed prompting improves \textit{Incorrect} prediction for GPT-5 and Qwen3-32B, whereas the benefit of providing the full study rather than selected passages is model-dependent. 
This suggests that additional evidence does not uniformly improve verdict prediction; models differ in how effectively they use broader scientific context.

Second, when models do reach the expected \textit{Incorrect} verdict, their reasoning paths do not necessarily match the human expert explanation (\emph{RQ2}). 
Table~\ref{tab:alignment_validation} shows that Qwen3-32B achieves the strongest verdict-level performance, while GPT-5 produces more fallacy-specific reasoning sub-graphs that align with the human expert path and are accepted by validation. 
This gap demonstrates that final-label accuracy and reasoning-path alignment capture distinct aspects of model behavior. 
A model can assign the expected verdict while relying on a different fallacy decomposition or evidence-use pattern than the human reference.

Third, when model reasoning diverges from the human path, the divergence is not always an error (\emph{RQ3}). 
The validation results show that many non-human-aligned sub-graphs remain grounded in the cited study, relevant to the claim, and sufficient to justify the verdict. 
These cases represent valid alternative reasoning paths rather than reasoning failures. 
At the same time, rejected sub-graphs expose genuinely defective reasoning, including unsupported, irrelevant, or insufficient justifications.

Overall, these findings show that scientific fact-checking evaluation should move beyond verdict accuracy alone. 
Typed reasoning graphs make it possible to distinguish four behaviorally important cases: human-aligned reasoning, valid alternative reasoning, ungrounded reasoning, and grounded but inadequate reasoning. 
This distinction is especially important for misleading scientific claims, where a correct \textit{Incorrect} label can mask substantial variation in how models interpret evidence, identify fallacies, and justify their decisions.

% The results answer our research questions by separating verdict accuracy from reasoning quality. 
% For RQ1, Table~\ref{tab:combined_results} shows that LLMs can identify many misleading claims as \textit{Incorrect}, but performance depends on the model, prompt, and evidence setting. 
% Detailed prompting improves \textit{Incorrect} prediction for GPT-5 and Qwen3-32B, while the effect of full-study versus selected-passage evidence is model-dependent.

% For RQ2, Table~\ref{tab:alignment_validation} shows that correct verdict prediction does not imply human-aligned reasoning. 
% Qwen3-32B achieves the strongest verdict-level performance, but GPT-5 produces more human-aligned and accepted reasoning sub-graphs. 
% Thus, final-label accuracy and reasoning-path alignment measure different aspects of model behavior.

% For RQ3, the validation results show that non-alignment with human reasoning is not always failure. 
% Many non-human-aligned sub-graphs are still accepted after grounding and relevance--sufficiency validation, indicating valid but different reasoning paths. 
% At the same time, rejected sub-graphs reveal cases where LLM reasoning is unsupported or insufficient.

% Overall, our findings show that scientific fact-checking evaluation should move beyond verdict accuracy. 
% Reasoning graphs make it possible to distinguish human-aligned reasoning, valid alternative reasoning, not grounded reasoning, and grounded but inadequate reasoning.
\section{Conclusion}

% We presented a graph-based framework for comparing human and LLM reasoning paths in scientific fact-checking. 
% Instead of evaluating only whether a model predicts the expected \textit{Incorrect} verdict, our approach represents explanations as reasoning graphs and compares them at the level of fallacy-specific sub-graphs. 
% For non-human-aligned paths, we further assess whether the reasoning is grounded in the cited study, relevant to the claim, and sufficient to justify the verdict.

% Experiments on 84 \textsc{MissciPlus} false claims show that verdict accuracy and reasoning-path quality are distinct evaluation dimensions. 
% Qwen3-32B achieves the strongest verdict-level performance, while GPT-5 shows stronger human-aligned reasoning; in addition, many non-human-aligned paths across models are accepted after validation. 
% These results indicate that LLMs may reach the expected verdict without following the same reasoning path as human experts, but that divergence does not necessarily imply invalid reasoning.
%By distinguishing human-aligned reasoning, valid alternative reasoning, not grounded reasoning, and grounded but inadequate reasoning, our framework provides a fine-grained view of how LLMs reason over cited scientific evidence.

We present a graph-based framework for evaluating LLM reasoning in scientific fact-checking by comparing typed reasoning graphs with human expert paths and validating divergent paths for grounding, relevance, and sufficiency. Our Experiments %on 84 \textsc{MissciPlus} false claims 
shows that verdict accuracy and reasoning quality diverge: Qwen3-32B performs best at verdict prediction, while GPT-5 produces more human-aligned reasoning, and many non-human-aligned paths remain valid alternatives. These results suggest that LLM-based systems have the potential to serve as explanation aids: they can help experts inspect candidate reasoning paths and help readers understand how claims misrepresent cited evidence. However, final judgments should remain human-led, since models can still produce reasoning paths that are unsupported, irrelevant, or insufficient, and distinguishing these failures from valid alternative reasoning often requires expert assessment.

\section*{Limitations}

We outline several limitations of the present study; addressing them is the focus of our planned follow-up work.

\paragraph{Sample size and scope.}
Our evaluation uses  84 false claims from the \textsc{MissciPlus} test set, which constrains the statistical resolution of cross-model comparisons. We do not report confidence intervals, paired bootstrap estimates over claims, or significance tests; differences of a few percentage points in Tables~\ref{tab:combined_results} and~\ref{tab:alignment_validation} should therefore be read as indicative rather than conclusive. The dataset is also restricted to English-language biomedical and health claims, so generalization to other languages and scientific domains remains open.

\paragraph{Human annotation protocol.}
Both the HealthFeedback gold extraction (Section~\ref{subsec:data-evidence}) and the human re-check of QA-based grounding (Section~\ref{subsec:validation}) were performed by a single annotator (an author of this paper), not blinded to the source LLM. We therefore did not measure inter-annotator agreement and did not perform double annotation with adjudication. The grounding re-check shifts the \emph{Grounded (final)} column by 5--19 percentage points relative to the QA-only baseline in Table~\ref{tab:alignment_validation}, so a stricter multi-annotator and blinded protocol---particularly for the re-check step---is a clear priority for future work.

\paragraph{Evaluation set design.}
Because all 84 instances are false claims, the reported \textit{Incorrect} rate corresponds to recall on the expected label rather than to verdict accuracy. The \textit{Correct} column reflects false-positive endorsement of misleading claims and \textit{NEI} reflects abstention. A model that systematically prefers \textit{Incorrect} can therefore score highly on this metric without engaging with the evidence; a held-out set of correctly-supported claims (e.g., drawn from HealthFC or SciFact) would be needed to compute precision on \textit{Incorrect} and to separate verdict-class bias from genuine fact-checking ability.

\paragraph{Training-data exposure.}
\textsc{MissciPlus} and the underlying HealthFeedback reviews are public web content that may have been seen during pretraining or post-training of the proprietary models (GPT-5, Claude Opus 4.7). The contamination check on ten 2025-2026 claims (Section~\ref{subsec:contamination-check}) is too small to bound this effect.
% and we did not perform verbatim-overlap or memorization probes against the gold reviews. 
Reported human-alignment rates for the proprietary models may therefore be inflated relative to Qwen3-32B, which is an open-weight model with weaker memorization of the same web sources.


\bibliography{custom}

\clearpage
\appendix

\section{Related Work}

%\subsection{Scientific and Biomedical Claim Verification}
\paragraph{Scientific and Biomedical Claim Verification.}

Scientific fact-checking is commonly treated  as verifying claims against scientific evidence. 
\texttt{SciFact} evaluates evidence retrieval, rationale selection, and veracity prediction \citep{wadden-etal-2020-fact}, while \texttt{PUBHEALTH}, \texttt{HEALTHVER}, and \texttt{HealthFC} focus on public-health and biomedical claim verification \citep{kotonya-toni-2020-explainable-automated,sarrouti-etal-2021-evidence-based,vladika-etal-2024-healthfc}. 
Later work extends this setting to open-domain and richer evidence scenarios, including \texttt{SciFact-Open}, \texttt{SciVer}, and \texttt{CLAIM-BENCH} \citep{wadden-etal-2022-scifact,wang-etal-2025-sciver,javaji-etal-2025-ai}. 
These benchmarks establish evidence-grounded claim verification, but they mainly evaluate verdicts, retrieved evidence, or rationale spans rather than the evidence-to-verdict reasoning path followed by the model.

%\subsection{Misrepresentation of Scientific Evidence}
\paragraph{Misrepresentation of Scientific Evidence.}

A closely related line of work studies how scientific evidence is misrepresented. 
\textsc{Missci} models misleading scientific claims as fallacious arguments, where real studies are used to support inaccurate claims through implicit reasoning errors \citep{glockner-etal-2024-missci}. 
\textsc{MissciPlus} grounds these fallacies in passages from the misrepresented publications \citep{glockner-etal-2025-grounding}. 
These studies show that scientific misinformation often arises from distorted reasoning over valid evidence. 
However, they primarily focus on reconstructing premises, grounding fallacies, or classifying fallacy types, rather than comparing human and LLM explanations as parallel reasoning paths.

\paragraph{LLMs for Fact-Checking and Explanation Generation.}

Recent work evaluates LLMs for scientific, biomedical, and public-health fact-checking. 
LLMs can generate plausible verdicts and explanations when guided by evidence or structured prompts \citep{alvarez-etal-2024-zero,liang-sonntag-2025-advancing,vladika-etal-2025-step,zarharan-etal-2024-tell,tan-etal-2025-improving}. 
However, explanation-faithfulness studies show that fluent explanations may not faithfully reflect the underlying decision process \citep{turpin-etal-2023-language,atanasova-etal-2023-faithfulness,parcalabescu-frank-2024-measuring}. 
This motivates evaluating whether LLM explanations follow expert reasoning paths, not only whether they provide plausible evidence-based justifications.

\paragraph{Human Rationales and Structured Reasoning Explanations.}

Our work also relates to human rationales and structured explanations. 
Datasets such as e-SNLI and ERASER use human explanations and rationales as evaluation targets beyond final labels \citep{camburu-etal-2018-snli,deyoung-etal-2020-eraser}. 
Explanation-graph and multi-step reasoning work, including WorldTree and EntailmentBank, represents explanations as connected reasoning structures \citep{jansen-etal-2018-worldtree,xie-etal-2020-worldtree,dalvi-etal-2021-explaining}. 
Argument-mining research similarly models claims, premises, warrants, and support relations \citep{lawrence-reed-2019-argument}. 
Building on these ideas, we represent fact-checking explanations as reasoning graphs and compare them at the level of fallacy-specific sub-graphs.

\paragraph{Grounding and Faithfulness Evaluation.}

Grounding and faithfulness work examines whether generated content is supported by source documents. 
Prior studies evaluate attribution, citation faithfulness, QA-based factuality, and atomic-fact support \citep{gao-etal-2023-enabling,min-etal-2023-factscore,fabbri-etal-2022-qafacteval,tang-etal-2024-minicheck,li-etal-2024-attributionbench}. 
These approaches usually assess grounding at the level of claims, sentences, citations, or atomic facts. 
Our study shifts the focus to grounded reasoning paths: for non-human-aligned LLM graphs, we test whether the alternative path is grounded in the cited study, relevant to the claim, and sufficient for the \textit{Incorrect} verdict.

These lines of work address complementary parts of process-level evaluation but remain disconnected. Verification benchmarks emphasize verdicts and retrieved evidence; fallacy work identifies reasoning errors; faithfulness and grounding work assess whether explanations are decision-consistent or evidence-supported; and structured-rationale datasets model generic reasoning chains rather than fallacy-specific paths. Our framework connects these perspectives by aligning human and LLM explanations through fallacy-specific subgraphs and validating non-aligned LLM paths for grounding in the cited study, relevance to the claim, and sufficiency for the verdict.

%These five lines of work---scientific claim verification, fallacy reconstruction, LLM-generated explanations and their faithfulness, structured-rationale and explanation-graph datasets, and grounding evaluation---each address one aspect of process-level evaluation but do not connect them. Verification benchmarks score the final verdict and possibly the retrieved evidence; fallacy work classifies the reasoning error in isolation; faithfulness work asks whether an explanation reflects the model's decision but not whether it follows the same path an expert would take; structured-rationale work compares explanations at the level of generic entailment trees rather than fallacy-specific reasoning units; and grounding work checks atomic-fact support rather than reasoning-chain support. Our framework bridges these by aligning human and LLM reasoning at the level of fallacy-specific sub-graphs and, for non-aligned LLM paths, validating grounding in the cited study, relevance to the claim, and sufficiency for the verdict in a single pipeline.

\begin{table}[t]
\centering
\scriptsize
\setlength{\tabcolsep}{4pt}
\resizebox{\columnwidth}{!}{%
\begin{tabular}{lccc}
\toprule
\textbf{Metric} & \textbf{Claude Opus 4.7} & \textbf{GPT-5} & \textbf{Qwen3-32B} \\
\midrule
\multicolumn{4}{l}{\textbf{Veracity}} \\
Correct & 30.0 & 10.0 & 0.0 \\
Incorrect & 70.0 & 70.0 & \textbf{80.0} \\
Not Enough Information & 0.0 & 20.0 & 20.0 \\
No Response & 0.0 & 0.0 & 0.0 \\
\midrule
Consistency & \textbf{90.0} & \textbf{90.0} & \textbf{90.0} \\
\midrule
Human Alignment & 27.8 & \textbf{64.7} & 11.8 \\
\bottomrule
\end{tabular}%
}
\caption{Contamination check on 10 false claims published in 2025--2026 that are not part of the dataset used in this study. Values are reported as \%.}
\label{tab:latest-claims-contamination}
\end{table}

\begin{figure*}[t]
  \centering
  \includegraphics[width=\textwidth]{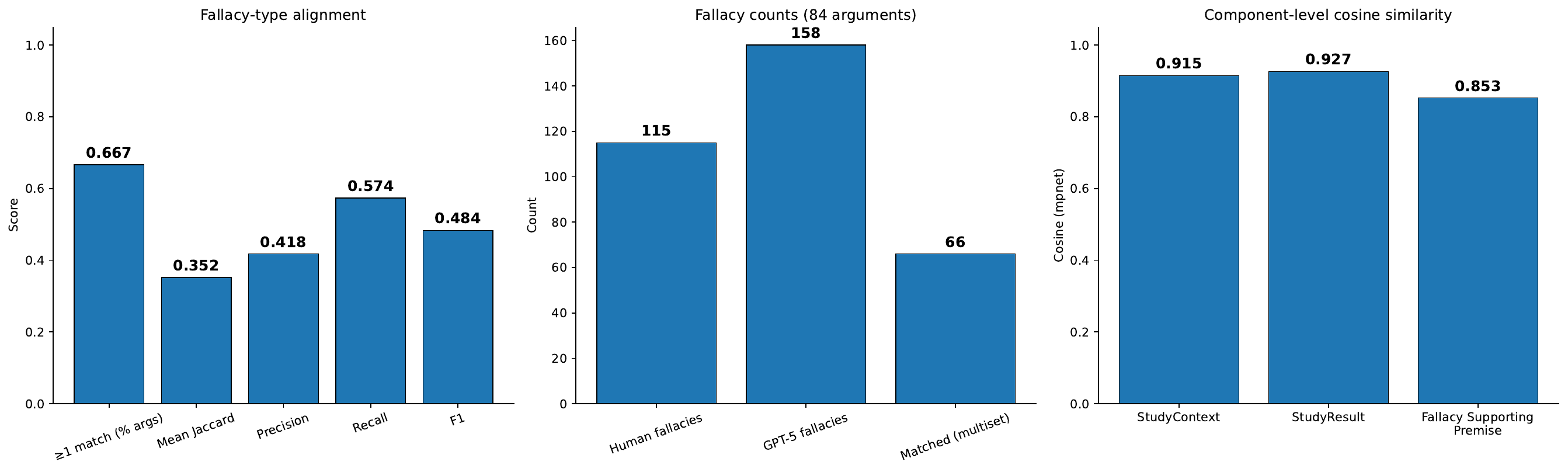}
  \caption{Agreement between human and GPT-5 annotations for the same 84 GPT-5 fact-checking explanations. Each explanation was independently converted into a reasoning graph by a human annotator and by GPT-5. The figure summarizes the extent to which the GPT-5-generated reasoning graphs align with the human reference graphs. Left: fallacy-type agreement at the argument level, including at-least-one type match, mean Jaccard similarity, precision, recall, and $F_1$ under type-equal multiset matching. Middle: total fallacy labels produced by each annotator and the subset matched between them. Right: mean cosine similarity using \texttt{all-mpnet-base-v2} for corresponding textual components: \textit{Study Context}, \textit{Study Result}, and \textit{Fallacy-Supporting Premise}. For fallacy-supporting premises, each GPT-5 premise is scored against its best-matching human premise and then averaged at the argument level.}
  \label{fig:human_vs_gpt5_alignment}
\end{figure*}

\section{Fallacy Abbreviations}
\label{app:fallacy-abbrev}

Table~\ref{tab:fallacy-abbrev} lists the abbreviations used along the
axes of the fallacy-distribution panels (Figure~\ref{fig:qa_fallacy_dist})
together with their full names. The nine categories follow the taxonomy
used throughout the annotation protocol.

\begin{table}[t]
  \centering
  \small
  \begin{tabular}{l l}
    \toprule
    \textbf{Abbr.} & \textbf{Full name} \\
    \midrule
    FOE & Fallacy of Exclusion \\
    CO  & Causal Oversimplification \\
    HG  & Hasty Generalization \\
    FE  & False Equivalence \\
    AM  & Ambiguity \\
    BSP & Biased Sample Fallacy \\
    FOC & Fallacy of Composition \\
    FD  & False Dilemma \\
    IE  & Impossible Expectations \\
    \bottomrule
  \end{tabular}
  \caption{Abbreviations for the nine fallacy categories used in
  Figure~\ref{fig:qa_fallacy_dist}.}
  \label{tab:fallacy-abbrev}
\end{table}

\begin{figure}[t]
  \centering
  \includegraphics[width=\columnwidth]{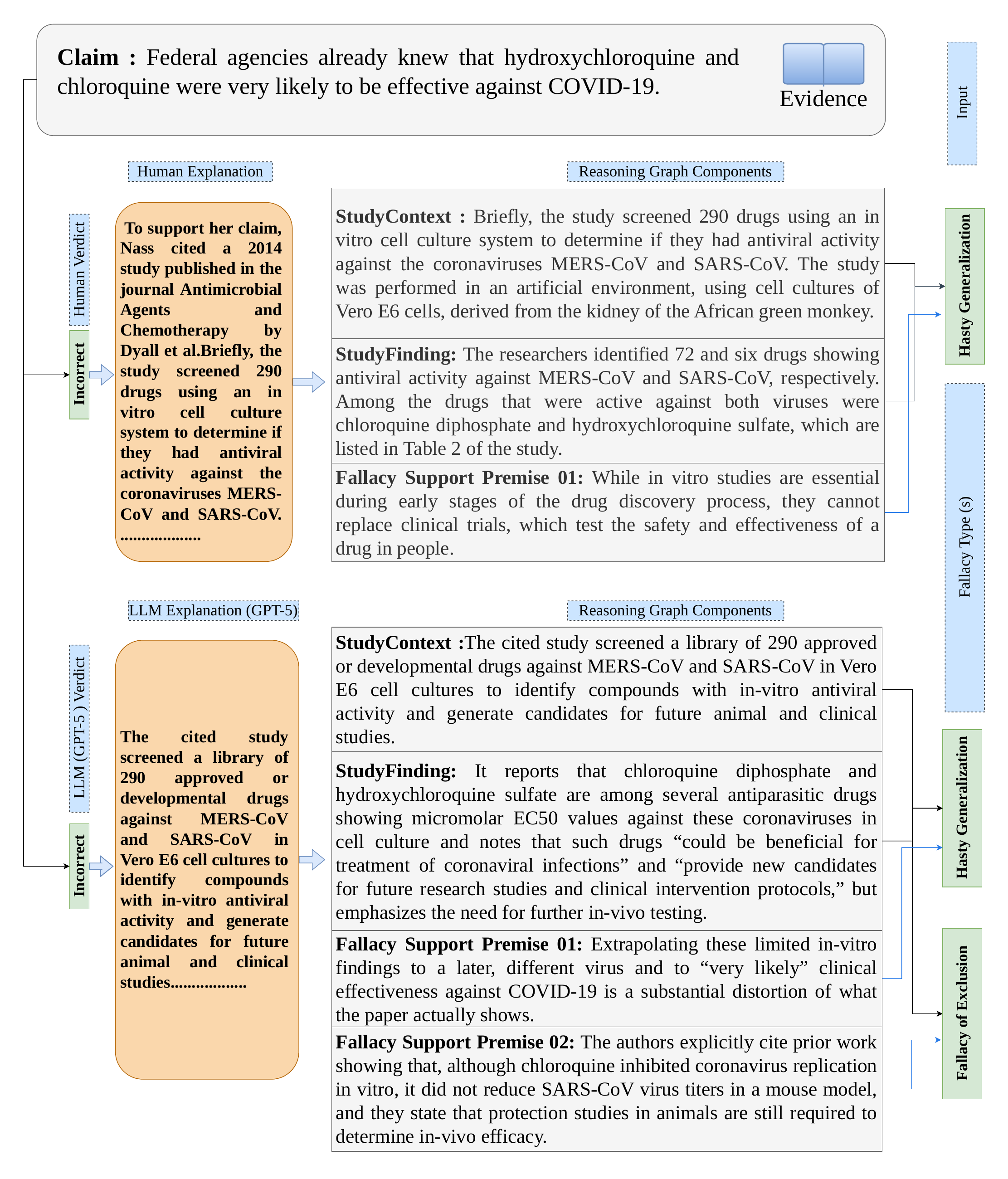}
  \caption{Illustration of human and LLM reasoning paths for a scientific fact-checking example. The example shows that an LLM reasoning path can align with the human reasoning path, while it can also take a different but valid reasoning path.}
  \label{fig:detailed_intro_diagram}
\end{figure}

\begin{figure*}[t]
\centering
\begin{minipage}{0.97\textwidth}
\begin{Verbatim}[
fontsize=\scriptsize,
breaklines=true,
breakanywhere=true,
frame=single,
framesep=3mm
]
You are an expert scientific fact-checker specializing in evaluating whether public claims
accurately represent findings from scientific research papers. You have access to the full cited
study provided as input. Your task is to determine if the claim faithfully reflects the evidence
presented in that study.

Input

## Claim:
{[claim]}

## Cited Study:
{[study]}

Instructions

1. Study Understanding
Carefully read and summarize the main purpose, methods, and findings of the cited study.
Identify the key conclusions or outcomes that the authors report.

2. Claim Interpretation
Summarize what the claim asserts or implies.
Identify whether the claim refers to a specific result, a general conclusion, or a causal statement.

3. Evidence Mapping
Locate and quote or paraphrase the specific parts of the study that:
(a) Support the claim, if any.
(b) Contradict or differ from what the claim asserts.
(c) Have been distorted or exaggerated by the claim.

4. Verdict
Choose one clear verdict:

Correct -- The claim correctly represents the cited study.
Incorrect -- The claim misrepresents or contradicts the study's findings.
Not Enough Information -- The study does not provide adequate evidence to support or refute the claim.

5. Reasoning behind verdict
Provide the justification and reasoning behind the final verdict in 3--10 sentences, based on the
evidence above.

6. Output Format
{Justification and reasoning behind final verdict in 3--10 sentences based on the evidence above.}
{"final_verdict": "Correct / Incorrect / Not Enough Information"}
\end{Verbatim}
\end{minipage}
\caption{Scientific fact-checking prompt used to assess whether a claim faithfully represents the evidence presented in its cited scientific paper.}
\label{fig:factcheck-prompt}
\end{figure*}

\begin{figure*}[t]
\centering
\begin{minipage}{0.97\textwidth}
\begin{Verbatim}[
fontsize=\scriptsize,
breaklines=true,
breakanywhere=true,
frame=single,
framesep=3mm
]
You are a careful annotator. Score the semantic textual similarity (STS) of two texts using the
rubric the user provides. Reply with valid JSON only.

Classify the following pair of arguments by SIMILARITY using this scale:

(5) Completely equivalent, meaning pretty much exactly the same thing, using different words.

(4) Mostly equivalent, but some unimportant details differ. One argument may be more specific
than another or include a relatively unimportant extra fact.

(3) Roughly equivalent, but some important information differs or is missing. This includes cases
where the argument is about the same FACET but the authors have different stances on that facet.

(2) Not equivalent, but share some details. For example, talking about the same entities but
making different arguments or addressing different facets.

(1) Not equivalent, but on the same topic.

(0) On a different topic.

Text A:
{human_text}

Text B:
{llm_text}

Return only valid JSON in this format:

{"sts": <integer from 0 to 5>}
\end{Verbatim}
\end{minipage}
\caption{Prompt used to score semantic textual similarity (STS) between human and LLM reasoning components. \citep{misra-etal-2015-using}}
\label{fig:sts-prompt}
\end{figure*}

\begin{figure*}[t]
\centering
\begin{minipage}{0.95\textwidth}
\begin{Verbatim}[
fontsize=\tiny,
baselinestretch=0.82,
breaklines=true,
breakanywhere=true,
frame=single,
framesep=1.2mm
]
You are an expert evaluator of scientific fact-checking explanations and argumentative reasoning.

Task: Evaluate whether the reasoning graph is relevant to the Claim and sufficient to justify an
Incorrect verdict.

Important: Assume the graph components are valid. Judge only whether they are relevant to the
Claim and sufficient to support the Incorrect verdict. The Claim is the target statement being
assessed; it is not itself evaluated as a graph component.

Target Statement

Claim:
{[CLAIM]}

Reasoning Graph Components

StudyContext:
{[STUDY_CONTEXT]}

StudyResult:
{[STUDY_RESULT]}

ContextPremises / FallacySupportingPremises:
{[CONTEXT_PREMISES]}

FallacyType, if available:
{[FALLACY_TYPE]}

FinalVerdict: Incorrect

Graph Component Definitions

1. StudyContext: Contextual information used in the graph, including scope, setting,
population, method, comparison, limitation, or other relevant details.

2. StudyResult: The study result, finding, conclusion, outcome, or reported effect represented
in the graph.

3. ContextPremises / FallacySupportingPremises: Reasoning premises that explain how the Claim
relates to the graph context and result, and why the Claim may be false, misleading, overstated,
unsupported, or inconsistent.

4. FallacyType, if available: The type of reasoning error identified in the explanation. Treat
this as an optional part of the reasoning chain, not the main evaluation target.

5. FinalVerdict: The target verdict for all samples in this task is Incorrect.

Evaluation Dimensions

1. Relevance to the Claim

Evaluate whether the graph components are relevant to the Claim.

Ask:
- Do the components address the same main assertion as the Claim?
- Are they focused on the Claim's key issue, entity, relation, population, outcome, comparison,
  or causal statement?
- Do they explain why the Claim may be false, misleading, overstated, unsupported, or inconsistent?
- Are any components off-topic or unrelated?

Score:
2 = Clearly relevant: the components directly address the Claim.
1 = Partially relevant: the components are related, but the connection is incomplete, indirect,
    or only partly focused.
0 = Not relevant: the components do not meaningfully address the Claim.

2. Sufficiency for the Incorrect Verdict

Evaluate whether the graph components, taken together, provide enough support for judging the
Claim as Incorrect.

Ask:
- Assuming the graph components are true, do they justify an Incorrect verdict?
- Does the chain explain why the Claim is false, misleading, overstated, unsupported, or
  inconsistent?
- Are important reasoning steps missing?
- Could the same components reasonably support a verdict other than Incorrect?

Score:
2 = Clearly sufficient: the graph chain supports an Incorrect verdict.
1 = Partially sufficient: the graph chain gives some support, but important reasoning is missing,
    weak, or underdeveloped.
0 = Not sufficient: the graph chain does not justify an Incorrect verdict.

Overall Adequacy:
- Adequate: relevance = 2 and sufficiency = 2.
- Partially adequate: neither relevance nor sufficiency is 0, but at least one score is 1.
- Not adequate: relevance = 0 or sufficiency = 0.

Important Instructions:
- Do not require the graph to match a human expert's reasoning path.
- Do not focus only on the FallacyType.
- Treat FallacyType only as an optional component in the reasoning chain.
- Give short, direct reasons.
- Return only valid JSON.
- Do not include markdown, explanations outside JSON, or extra text.

Return format:

{
  "relevance_to_claim": {
    "score": 0,
    "label": "Clearly relevant / Partially relevant / Not relevant",
    "reason": ""
  },
  "sufficiency_for_incorrect_verdict": {
    "score": 0,
    "label": "Clearly sufficient / Partially sufficient / Not sufficient",
    "reason": ""
  },
  "overall_adequacy": {
    "label": "Adequate / Partially adequate / Not adequate",
    "reason": ""
  }
}
\end{Verbatim}
\end{minipage}
\vspace{-1mm}
\caption{Prompt used to evaluate whether non-human-aligned reasoning sub-graphs are relevant to the claim and sufficient to justify an \textit{Incorrect} verdict.}
\label{fig:relevance-sufficiency-prompt}
\end{figure*}

\begin{figure*}[t]
\centering
\begin{minipage}{0.95\textwidth}
\begin{Verbatim}[
fontsize=\tiny,
baselinestretch=0.82,
breaklines=true,
breakanywhere=true,
frame=single,
framesep=1.2mm
]
You are converting a fact-checking explanation into a structured reasoning graph and assigning
fallacy labels.

You will receive a CLAIM and an EXPLANATION that evaluates the claim using a primary scientific
study.

VERBATIM EXTRACTION RULE: All textual fields below (StudyContext, StudyResult,
supporting_premise) MUST be COPIED VERBATIM from the EXPLANATION. Do NOT paraphrase,
summarise, rephrase, abbreviate, expand, or change the wording in any way. Only select
word-for-word spans from the explanation. Preserve original wording, punctuation, numbers,
casing, and references exactly as written.

Reasoning flow: perform internally IN THIS ORDER, then emit the JSON.

Step 1 -- Find StudyContext
Locate the span(s) in the explanation that describe the cited primary study: study design,
population, setting, time period, and what was measured. Take the relevant sentence(s)
verbatim. If multiple studies are referenced, focus on the primary study the claim is based on.
If the explanation does not contain study context, leave StudyContext as an empty string.

Step 2 -- Find StudyResult
Locate the span(s) in the explanation that state the study's reported finding(s). Take them
verbatim. Numbers and effect sizes must be preserved exactly. If the explanation does not state
a study finding, leave StudyResult as an empty string.

Step 3 -- Collect candidate supporting premises
From the explanation, pick out single sentences that help REFUTE the claim, e.g., sentences that
point out biases, scope limits, missing evidence, oversimplifications, ambiguity, or other
reasons the claim does not follow from the study. Each candidate must be one sentence copied
verbatim from the explanation. Ignore background, tangential, or non-refuting sentences.

Step 4 -- Pair-and-classify
For each candidate supporting_premise, pair it with StudyContext + StudyResult and ask whether
this 3-tuple identifies one of the allowed fallacy types. If yes, emit a fallacy entry whose
`type` is that fallacy and whose `supporting_premise` is that single verbatim sentence. If the
3-tuple does not clearly fit any allowed fallacy type, drop that supporting_premise. Even if
StudyContext or StudyResult is empty, classify a fallacy using whatever information is available.

Step 5 -- Empty Fallacies when the claim is supported
If the explanation indicates the claim is correct or well-supported by the study, return an empty
Fallacies list. Still emit StudyContext and StudyResult.

Allowed fallacy types:
[
  "Ambiguity", "Impossible Expectations", "False Equivalence", "False Dilemma",
  "Biased Sample Fallacy", "Hasty Generalization", "Causal Oversimplification",
  "Fallacy of Composition", "Fallacy of Exclusion"
]

Fallacy Definitions:
[definitions and logical forms for each of the nine fallacy types follow the MISSCI taxonomy;
abbreviated here for space, full text in the released code.]

Input

Claim:
<CLAIM TEXT>

Explanation:
<EXPLANATION TEXT>

Output requirements
- Return only valid JSON. No prose, no code fences.
- Schema: exactly these keys, in this order.
{
  "StudyContext": "...",
  "StudyResult": "...",
  "Fallacies": [
    {
      "type": "<one of the allowed fallacy types>",
      "supporting_premise": "<single most-relevant premise sentence>"
    }
  ]
}
- supporting_premise must be a non-empty string: a single sentence taken VERBATIM from the
  explanation. It is NOT a list and NOT an index.
- Only emit a fallacy entry when the 3-tuple clearly identifies one of the allowed fallacy types.
- StudyContext, StudyResult, and supporting_premise text MUST appear word-for-word in the
  EXPLANATION. Do not rewrite or summarise.
- StudyContext or StudyResult may be empty strings if the explanation does not contain that
  information.
- "Fallacies" must be an empty list when the explanation supports the claim.
\end{Verbatim}
\end{minipage}
\vspace{-1mm}
\caption{Prompt used to extract structured reasoning graphs from free-text fact-checking explanations. The prompt is applied to both human expert explanations and LLM-generated explanations. The verbatim-extraction rule constrains the extractor to copy spans word-for-word. The full fallacy definitions block, omitted here for space, follows the nine-class MISSCI taxonomy \citep{glockner-etal-2024-missci}.}
\label{fig:graph-extraction-prompt}
\end{figure*}

\begin{figure*}[t]
\centering
\begin{minipage}{0.97\textwidth}
\begin{Verbatim}[
fontsize=\scriptsize,
breaklines=true,
breakanywhere=true,
frame=single,
framesep=3mm
]
You are helping with premise verification.

<premise label>:
<premise text>

Generate exactly 5 short factual questions from this premise.

Rules:
- Each question must be directly based on a specific factual statement in the premise.
- The answer to each question must be contained in the premise itself.
- Do not ask broad, random, background, or speculative questions.
- Do not ask questions that require information outside the premise.
- Questions must focus on the core factual content needed to verify the premise against the
  scientific paper.
- Return only valid JSON in this format:

{
  "questions": ["...", "...", "...", "...", "..."]
}
\end{Verbatim}
\end{minipage}
\caption{Prompt used with GPT-5 to generate five factual questions per premise during the QA-grounding stage. \texttt{<premise label>} is one of \emph{Study context premise}, \emph{Study finding premise}, or \emph{Fallacy supporting premise}.}
\label{fig:prompt-qa-question-gen}
\end{figure*}
\end{document}